\documentclass[letterpaper, 10 pt, conference]{ieeeconf}  

\IEEEoverridecommandlockouts                              

\usepackage{xcolor}
\usepackage{xspace}
\usepackage{graphicx}
\usepackage{booktabs} 
\usepackage{tabularx}
\usepackage{multirow}
\usepackage[table]{xcolor}
\usepackage{caption}
\usepackage{dblfloatfix}
\usepackage[hidelinks]{hyperref}
\usepackage[many]{tcolorbox}
\newcolumntype{Y}{>{\centering\arraybackslash}X}
\usepackage{cuted}

\newcommand{\methodacro}{SAIL\xspace} 

\title{\LARGE \bf
SAIL: Test-Time Scaling for In-Context Imitation Learning with VLM
}

\author{
Makoto Sato$^{1}$, Yusuke Iwasawa$^{1}$, Yujin Tang$^{2}$, and So Kuroki$^{2}$%
\thanks{$^{\ast}$This work was done while Makoto Sato was an intern at Sakana AI.}
\thanks{$^{1}$The University of Tokyo, Tokyo, Japan.}%
\thanks{$^{2}$Sakana AI, Tokyo, Japan.}%
}

\begin{document}
\maketitle


\thispagestyle{empty}
\pagestyle{empty}

\begin{strip}
\vspace{-7em}
\begin{center}
  \begin{minipage}{\textwidth}
    \centering
    \includegraphics[width=\linewidth]{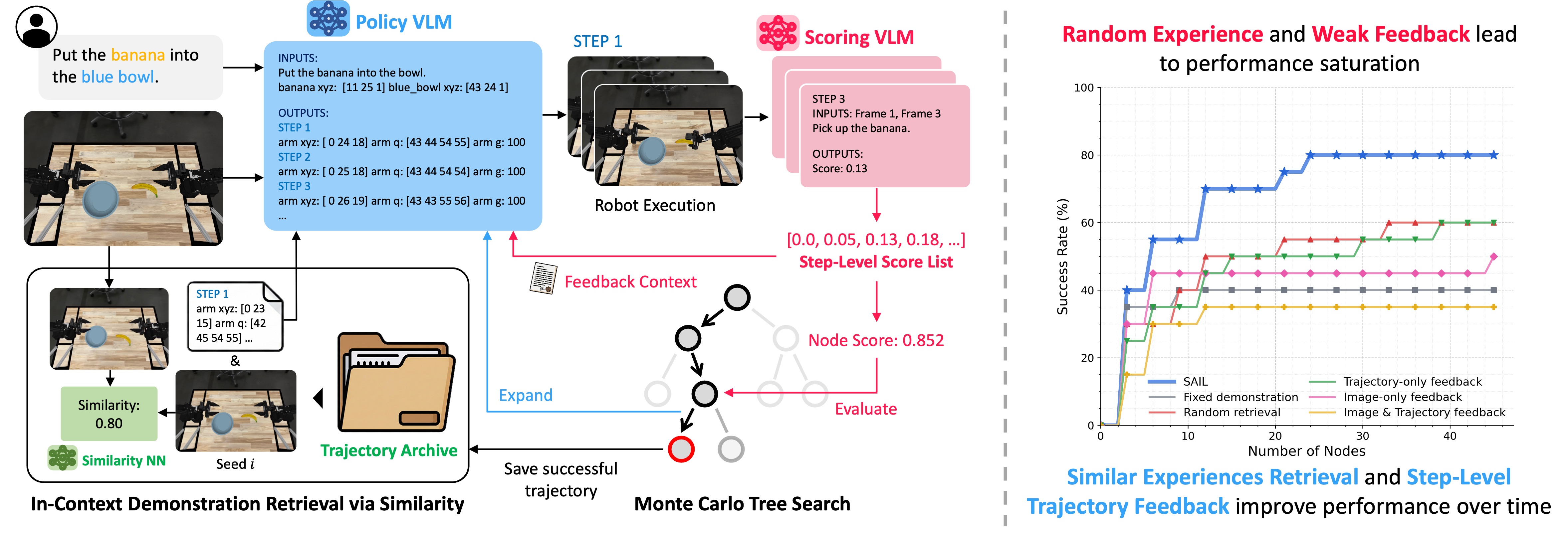}
    \vspace{-2em}
    \captionof{figure}{\textnormal{\textbf{Left:} We perform test-time scaling with MCTS over VLM-proposed trajectories, using similarity-based retrieval from an archive of successful rollouts and step-level feedback from a VLM evaluator. \textbf{Right:} Success improves with more expanded MCTS nodes (test-time compute); random retrieval and weak feedback saturate early, whereas our retrieval and step-level feedback keep yielding gains.}}
    \label{fig:scaling_curve}
    \vspace{-1em}
  \end{minipage}
\end{center}
\end{strip}


\begin{abstract}
In-context imitation learning allows robots to acquire skills from demonstrations, yet one-shot trajectory generation remains fragile under environmental variation.
We propose \methodacro, a framework that reframes robot imitation as an iterative refinement problem capable of scaling with test-time compute.
\methodacro utilizes Monte Carlo Tree Search, where each node is a complete trajectory and edges correspond to trajectory refinements.
The process is guided by three core components:
an automated archive of successful trajectories for contextually relevant retrieval, a vision language model-based scoring mechanism for trajectory evaluation, and a step-level feedback that provides trajectory-aligned scores for iterative refinement.
Experiments across six diverse manipulation tasks in simulation and real-world validation clearly demonstrate that increasing test-time compute consistently improves success rates, achieving up to 95\% on complex tasks.
Our results suggest that trajectory-level test-time scaling is a robust path toward more generalizable robotic agents.
\end{abstract}


\section{Introduction}
\label{sec:intro}
In-context learning has emerged as a compelling paradigm for robots to acquire complex skills directly from a few visual hints and trajectory demonstrations~\cite{shah2025mimicdroid,dipalo2024kat,fu2024context}.
However, a fundamental bottleneck remains because a single realized trajectory often fails to generalize when the robot encounters new initial conditions.
While current vision language models (VLMs) can generate trajectories~\cite{zitkovich2023rt,puthumanaillam2025trace}, they typically treat this as a one-shot prediction task.
This limitation means that the success of the system is capped by the initial guess of the model, which cannot adjust for environmental variations during the inference pass.

In this work, we propose that robot imitation should instead be framed as an iterative search problem where performance can be improved by scaling test-time compute.
By treating full trajectories as the objects of search, we enable the system to explore potential solutions in a trial-matched simulator before execution.
This approach shifts the focus from static imitation to a dynamic paradigm where a robot can effectively think longer to resolve the ambiguities of a novel task.

The reliance on foundation models' one-shot inference is problematic because the high dimensional space of robot trajectories is extremely sensitive to small errors in initial state estimation or object localization.
Even a slight misalignment in the predicted end effector pose can lead to cascading failure during execution which the model cannot correct without an external loop for evaluation.
Recent efforts have attempted to mitigate this fragility by either distilling procedural reasoning into more robust visual representations~\cite{gu2023rt,lv2024robomp}, or employing retrieval mechanisms to provide the model with richer context from successful historical demonstrations~\cite{kuang2024ram}.
However, these methods typically focus on improving the quality of a single prediction rather than enabling the system to systematically explore and refine the continuous motion space itself.
While some frameworks incorporate external validators to guide corrections~\cite{du2023successdetectors,sontakke2023roboclip}, they often remain limited to symbolic planning and do not address the fundamental need for test-time scaling at the level of full robot trajectories.

To address these challenges we introduce \methodacro (\textbf{S}c\textbf{a}ling \textbf{I}n-context Imitation \textbf{L}earning), a novel framework that enables in-context imitation to scale through test-time compute (see Figure~\ref{fig:scaling_curve} left). 
Rather than relying on a VLM's single prediction, \methodacro treats trajectory generation as a refinement problem, and introduces three key components to guide the refinement process toward success.
First, we formulate trajectory generation in a Monte Carlo Tree Search (MCTS) setting where each node represents a complete robot trajectory.
This enables the system to systematically explore the continuous motion space rather than depending on a single one-shot plan.
Second, we maintain an automated archive of successful trajectories which are retrieved to provide contextually relevant demonstrations for new environmental configurations.
This retrieval mechanism allows the system to bootstrap its search by leveraging past experiences in visually similar scenes.
Finally, we propose a step-level feedback that uses trajectory-aligned scores to annotate trajectories.
This allows the generator to identify specific points of failure and iteratively refine its motion based on granular critiques.

We evaluate \methodacro across six diverse manipulation tasks in simulation and validate the full pipeline on a real robot system.
Our results demonstrate a clear scaling property where increasing the test-time compute budget consistently leads to higher success rates across all tasks (see Figure~\ref{fig:scaling_curve} right).
Notably, the system achieves a 95\% success rate on the HandOverBanana task and successfully transfers to the physical world with a high success rate on a block moving task using a trial matched digital twin.
These findings suggest that scaling search at the trajectory level is a viable path toward more robust and generalizable robotic agents.

Our contributions can be summarized as follows:
(1) We reformulate in-context robot imitation as a trajectory-level test-time refinement problem, enabling performance to scale beyond one-shot trajectory prediction.
(2) To realize this formulation, we develop \methodacro that combines trajectory-level MCTS, retrieval-augmented demonstrations, and step-level evaluation from a VLM for iterative refinement.
(3) We demonstrate through extensive experiments in simulation and the real world that increasing test-time compute leads to consistent improvements in task success rates.
\section{Related Work}
\label{sec:related_works}
\subsection{Trajectory Generation Using LLMs and VLMs}
\label{sec:related_traj_llm}
Recent work has demonstrated that large language models (LLMs) and VLMs can be used not only for high-level planning but also for direct generation of executable robot trajectories.
Early approaches show that an off-the-shelf LLM can generate a complete trajectory in a single forward pass, producing dense pose sequences that are directly executable in simulation or on real robots~\cite{kwon2024zeroshottraj}.
This direct but unconstrained generation is often brittle, leading to subsequent work that introduces more structured representations to improve robustness, such as tokenizing observations and actions or converting demonstrations into structured textual templates for in-context trajectory or action prediction~\cite{dipalo2024kat,yin2025roboprompt}.
Related efforts further emphasize structural conditioning by moving from dense trajectory generation to waypoint-level representations explicitly constrained by object kinematics~\cite{xia2024kinematicprompt}.
In parallel, trajectory editing methods address similar robustness issues by modifying an existing trajectory using language-derived waypoint updates or geometric constraints rather than generating a new trajectory from scratch~\cite{maurya2025trajadapt,huang2025gelato}.
Overall, these studies establish that direct trajectory generation with foundation models is feasible and can be stabilized through representation design and output-space constraints; however, trajectory generation is still typically performed as a one-shot or lightly iterative process for a single environment instance, and systematic test-time scaling across many initial conditions remains largely unexplored, which this work addresses.

\subsection{Planning and Test-Time Scaling}
\label{sec:related_planning_scaling}
Recent work in robot planning increasingly combines LLMs with external evaluation modules and search processes, rather than treating LLMs as standalone plan generators.
In this setting, planning is formulated as an iterative process in which generated plans are examined for feasibility and revised based on detected failures or inconsistencies, instead of being finalized in a single forward pass~\cite{curtis2024proc3s,wang2024llm3}.
For long-horizon tasks, verifying plans through constraint satisfaction or motion planning and incorporating the resulting feedback into subsequent generations has been shown to substantially improve executability~\cite{curtis2024proc3s}.
Beyond natural language plans, several studies propose generating plans in structured and verifiable intermediate representations, such as symbolic code or constrained programs, enabling external validators to guide corrections and framing planning as an optimization process~\cite{chen2025code_symbolic_planner,chi2025instructflow}.
Retrieval-based designs further extend these approaches by reusing past successful experiences at inference time, allowing adaptation to new environments through in-context retrieval~\cite{sridhar2025regent}.
Building on these foundations, several works explore test-time scaling by incorporating external scoring functions or search mechanisms, such as candidate selection based on external evaluations~\cite{ahn2022saycan}, systematic expansion of candidate solutions via MCTS guided by LLM outputs~\cite{lee2025prime_search}, and iterative replanning supported by structured environment representations~\cite{rana2023sayplan}.
While these approaches improve robustness and scalability of planning, they primarily focus on task-level decisions or planning structures rather than continuous motion generation.
In contrast, our work targets test-time scaling at the level of continuous robot trajectory generation itself, treating full trajectories as the objects of search and evaluation.

\subsection{Scoring Using VLMs}
Research on evaluating robot executions as videos has expanded with the development of VLMs, moving from success detection toward continuous scoring that captures the execution process. 
Early work used VLMs to judge success or failure by checking the consistency between post-execution observations and natural language task descriptions, showing that robot execution videos can serve as direct evaluation targets~\cite{du2023successdetectors,sontakke2023roboclip}. 
Subsequent studies considered video sequences as input and estimated task progress or value along the temporal dimension. 
For instance, Generative Value Learners (GVL) demonstrated that presenting frame sequences in context and reasoning over their temporal order can produce scores related to progress and value during execution~\cite{ma2024gvl}. 
Other VLM-based reward and evaluation models explicitly focus on manipulation processes and stage structures, including work on stage recognition for long-horizon manipulation, systematic studies of vision-language reward models for robot execution videos, and process-level evaluation of manipulation trajectories~\cite{chen2025sarm,lee2026roboreward,tan2025robodopamine}. 
In addition, several studies use VLMs to detect planning or execution errors from execution videos and to classify failure types, treating video evaluation as a diagnostic signal rather than a binary outcome~\cite{pacaud2025guardian,duan2024aha}. 
Across these lines of work, an important question is not only how to assign a single scalar score to an entire video, but also how to extract and use information along the execution timeline at an appropriate temporal resolution. 
In this context, ROVER addresses long robot execution videos by explicitly modeling temporal structure, recursively segmenting videos into subtask-aligned intervals and assigning scores at the timestep level~\cite{schroeder2025rover}. 
Following this trend, we use the VLM-based scorer to obtain both a trajectory-level score for search and timestep-level scores for feedback during robot trajectory generation.
\section{Background}
\label{sec:pf}
\subsection{Problem Formulation}
We consider a target manipulation task (e.g., handing over a banana or a pen) under multiple initial environmental conditions, indexed by a random seed. Let \textbf{$\omega\in\Omega$} denote a seed, and let \textbf{$\mathbf{I}_\omega$} be the initial RGB observation for seed $\omega$. For a small subset of seeds \textbf{$\Omega^{\prime}\subset\Omega$}, we assume access to one successful trajectory demonstration per seed, denoted by \textbf{$\tau_{\omega \in \Omega^{\prime}}$}.
Our goal is to generate a successful robot trajectory \textbf{$\tau_{\omega \in\Omega\setminus\Omega^{\prime}}$} for an unseen seed by leveraging these initial demonstrations.

The trajectory space is defined over the robot end-effector state \textbf{$\mathbf{e}_{\omega}$}, represented as a tuple of Cartesian position \textbf{$\mathbf{p}_{\omega}\in\mathcal{R}^3$}, orientation quaternion \textbf{$\mathbf{q}_{\omega}\in\mathcal{R}^4$}, and gripper status \textbf{$g_{\omega}\in\{\mathrm{OPEN},\mathrm{CLOSE}\}$}.
A complete trajectory for seed \textbf{$\omega$} is a sequence \textbf{$\tau_\omega=(\mathbf{e}_{\omega}^{(1)},\dots,\mathbf{e}_{\omega}^{(T)})$}.

To enable in-context imitation beyond one-shot prediction, we formulate inference as an iterative refinement process that runs independently per seed.
For a fixed seed \textbf{$\omega$}, at refinement step \textbf{$v$}\footnote{Please notice that $t\in[1,T]$ indicates the trajectory execution steps, and $v \in [1, V]$ is the test-time scaling refinement iteration.}, a policy VLM proposes a \emph{complete} trajectory \textbf{$\tau_\omega^{(v)}$} conditioned on
(1) the initial state \textbf{$\mathbf{s}_{\omega}$},
(2) a set of in-context reference demonstrations $\{\tau^{(v)}_{\omega_k}\}_{k=1}^{K}$ inserted into the prompt at step $v$,
and (3) feedback derived from previously executed trajectories along the current search path.
The policy VLM then generates \textbf{$\tau_\omega^{(v)}$}, which is then executed and evaluated to drive subsequent refinements, until a successful trajectory \textbf{$\tau_\omega^\star$} is obtained.

To construct the initial state representation \textbf{$\mathbf{s}_{\omega}$}, we extract task-relevant objects' spatial information from the initial RGB observation \textbf{$\mathbf{I}_\omega$}.
Concretely, a detection VLM predicts 2D keypoints for task-relevant objects, which are lifted to 3D world coordinates using calibrated camera parameters.
We form \textbf{$\mathbf{s}_{\omega}$} by concatenating the 3D keypoints with the initial robot state \textbf{$\mathbf{e}_{\omega}$}.

\subsection{Monte Carlo Tree Search}
We formulate trajectory refinement as a MCTS problem, where each node corresponds to a complete trajectory and edges correspond to refinement operations that modify a previous trajectory.
For each evaluation seed \textbf{$\omega\in\Omega$}, we build an independent search tree.
Each node corresponds to a complete trajectory proposal \textbf{$\tau_\omega^{(v)}$}, generated by the policy VLM conditioned on the initial state \textbf{$\mathbf{s}_{\omega}$} and a node-specific in-context demonstration set.
We execute each proposal in simulation and obtain an evaluation signal.

We use a standard MCTS loop consisting of \emph{selection, expansion, evaluation,} and \emph{backup}. For a node $n$, let $N(n)$ denote its visit count and let $\bar{R}(n)$ denote its empirical mean value. During selection, starting from the root, we repeatedly choose a child $c\in\mathrm{Ch}(n)$ that maximizes the Upper Confidence Bound (UCB) score, which balances exploitation and exploration. In this work, we adopt a prior-weighted UCB (PUCB~\cite{rosin2011multi}) variant that incorporates a node-specific prior to further enhance exploration toward promising branches:
\begin{equation}
c^\star = \operatorname*{arg\,max}_{c\in\mathrm{Ch}(n)}
\left(
\bar{R}(c) + c_{\mathrm{pucb}}*P(c)\sqrt{\frac{\ln N(n)}{N(c)}}
\right),
\end{equation}
where $c_{\mathrm{pucb}}>0$ is the exploration coefficient and $P(c)$ denotes a prior term proportional to the node’s score. During expansion, we generate \textbf{$B$} child nodes from the selected leaf by sampling \textbf{$B$} refined trajectories from the policy VLM (we refer to \textbf{$B$} as the \emph{branching factor}). Each child trajectory is executed and scored to obtain a scalar reward (Sec.~\ref{sec:method_score}), and these rewards are then backed up along the selected path to update $N(\cdot)$ and $\bar{R}(\cdot)$. We terminate when the test-time compute budget is exhausted (or early-stop once a successful trajectory is found) and return the best trajectory encountered.

\section{Methods}
\label{sec:methods}
\begin{figure*}[t]
    \centering
    \includegraphics[width=0.95\textwidth]{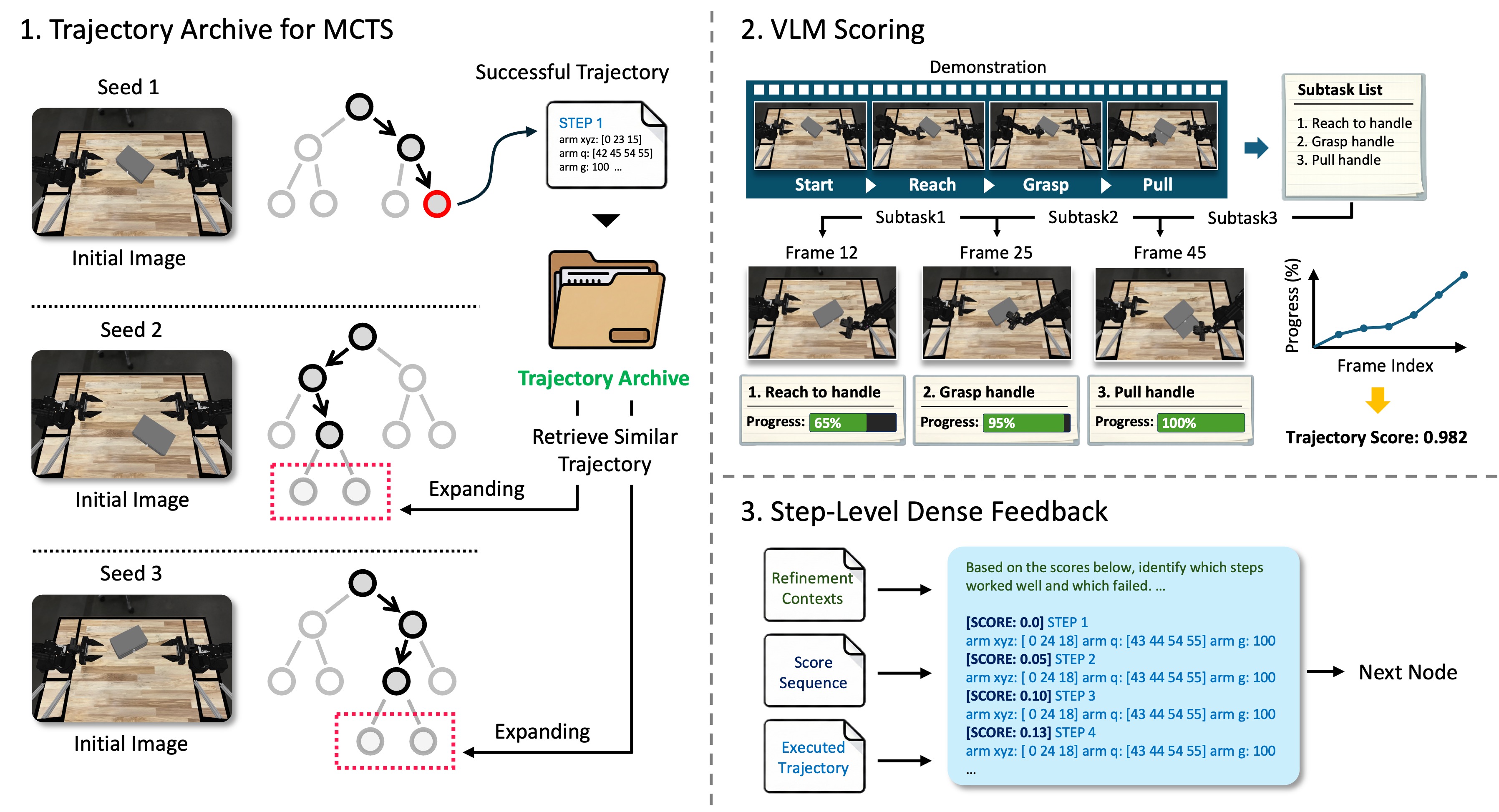}
    \caption{\textbf{Method overview.} \methodacro refines trajectories at test time via MCTS, where each node is a complete trajectory proposed by a policy VLM for a seed. (1) A shared \emph{trajectory archive} stores successful rollouts across seeds and retrieves visually similar trajectories as in-context demonstrations. (2) Each proposal is executed and scored by a scoring VLM that (i) decomposes the task into ordered subtasks from one demo and (ii) estimates per-frame completion to yield a scalar node value. (3) The scoring VLM aligns the progress scores to waypoints to provide step-level feedback for the next refinement.}
    \label{fig:cover_image}
\end{figure*}

\methodacro iteratively proposes, evaluates, and refines complete trajectories in a simulator. Figure~\ref{fig:cover_image} shows the method overview.
At a high-level, the policy VLM receives a discretized numeric encoding of \textbf{$\mathbf{s}_{\omega}$}, together with the in-context reference demonstrations and feedback from previous attempts, and outputs a candidate trajectory \textbf{$\tau_\omega^{(v)}$}.
An inverse kinematics (IK) controller executes the waypoints sequentially.
Then, a scoring mechanism evaluates the resulting rollout and returns both a scalar node value and step-level signals, which are used to guide iterative refinement in the next step.

In terms of test-time scaling for trajectory refinement, \methodacro consists of the following components that allows it to reuse successful trajectories across seeds, evaluate each candidate to guide node selection, and return dense feedback to refine subsequent proposals.

\begin{enumerate}
\item \textbf{Archive Retrieval for In-Context Demonstrations} (Sec.~\ref{sec:method_archive}): We maintain a shared archive of successful trajectories across seeds and retrieve visually similar trajectories for each node.
\item \textbf{Trajectory Scoring for Node Evaluation} (Sec.~\ref{sec:method_score}): We execute a generated trajectory in a simulator and use a scoring VLM to compute a scalar reward as the node value. 
\item \textbf{Step-Level Feedback for Trajectory Refinement} (Sec.~\ref{sec:method_fb}): We compute a progress signal from uniformly sampled rollout video frames and align these scores to the corresponding waypoints to provide dense feedback for the next refinement.
\end{enumerate}

\subsection{Archive Retrieval for In-Context Demonstrations}
\label{sec:method_archive}
In in-context imitation learning, performance depends strongly on how well the provided demonstrations match the test-time scene. We therefore maintain an \emph{archive} of successful trajectories that grows as the system searches across seeds, enabling retrieval of scene-compatible references over time. We define the archive as a set of successful rollouts:
\begin{equation}
\mathcal{A}=\{(\mathbf{I}_i,\tau_i)\}_{i=1}^{|\mathcal{A}|},
\end{equation}
where $\mathbf{I}_i \in \mathcal{R}^{H\times W\times 3}$ is the initial RGB observation and $\tau_i$ is the corresponding successful trajectory. We initialize the archive with the provided demonstrations across seeds, i.e., $(\mathbf{I}_\omega,\tau_{\omega})\in\mathcal{A}$ for each $\omega \in \Omega^{\prime}$. During search, whenever we discover a successful trajectory for a seed \textbf{$\omega$}, we append \textbf{$(\mathbf{I}_\omega,\tau_\omega^\star)$} to \textbf{$\mathcal{A}$}. As a result, \textbf{$\mathcal{A}$} contains successful trajectories originating from multiple seeds, and can be reused when searching for other seeds.

When expanding a node for a target seed \textbf{$\omega$} with initial observation \textbf{$\mathbf{I}_\omega$}, we retrieve a small set of visually similar archive entries to form the in-context examples.
We compute perceptual similarity using the LPIPS distance~\cite{zhang2018perceptual} and select the $K$ nearest neighbors by minimizing the total distance:
\begin{equation}
\mathcal{S}^\star(\mathbf{I}_\omega) = \operatorname*{arg\,min}_{\substack{S \subseteq \{1,\dots,|\mathcal{A}|\} \\ |S|=K}} \sum_{i\in S} \mathrm{LPIPS}(\mathbf{I}_\omega, \mathbf{I}_i).
\end{equation}

Let $\mathcal{S}^\star(\mathbf{I}_\omega)=\{i_1,\dots,i_K\}$.
We define $\tau^{(v)}_{\omega_k} := \tau_{i_k}$ for $k=1,\dots,K$,
and use $\{\tau^{(v)}_{\omega_k}\}_{k=1}^{K}$ as in-context references.

\subsection{Trajectory Scoring for Node Evaluation}
\label{sec:method_score}
Designing task-specific reward functions across diverse manipulation tasks is labor-intensive and does not scale.
We adopt a general VLM-based scoring mechanism that evaluates the executed rollout video and outputs a scalar reward for search.

A binary success signal is often too sparse for long-horizon tasks.
Following the trend of assigning timestep-level scores to capture the execution process~\cite{schroeder2025rover}, we evaluate progress over an ordered sequence of intermediate subtasks (e.g., reach, grasp, lift, handover in a hand over pen task).
To obtain these subtasks, we pre-generate a temporally ordered subtask list once per task using one available demonstration \textbf{$\tau_\omega$} from a seed \textbf{$\omega\in\Omega^{\prime}$}.
Specifically, we provide the scoring VLM with the final goal specification and the rollout video of \textbf{$\tau_\omega$}, and prompt it to decompose the task into $M$ ordered natural-language subtasks $\{\ell_m\}_{m=1}^{M}$.

Given $\{\ell_m\}_{m=1}^{M}$, we score a candidate trajectory for seed \textbf{$\omega$} by executing \textbf{$\tau_\omega^{(v)}$} in simulation and querying the VLM sequentially to estimate completion progress of the \emph{current} subtask.
For each subtask $\ell_m$, we treat the first frame of that subtask as a start reference (0\% completion) and evaluate subsequent frames in temporal order.
We uniformly sample $N$ rollout video frames along the temporal axis of the executed trajectory. 
At each evaluated rollout video frame $f \in [1, N]$, we query the VLM with the current subtask, the start reference frame, and the current frame, and parse its output into a completion percentage $\mathrm{VLM}(l_m, \tau_\omega^{(v)},f)\in[0,100]$ (i.e., we prompt the VLM to directly output the completion percentage).
When $\mathrm{VLM}(l_m, \tau_\omega^{(v)},f)$ reaches 100\%, we mark the subtask complete and proceed to $\ell_{m+1}$. If the rollout ends before completion, we stop and treat the trajectory as failing at that subtask.

We convert this subtask-level progress into a per-frame progress score \textbf{$r_\omega^{(v)}(f)\in[0,1]$}:
\begin{equation}
r_\omega^{(v)}(f)=\frac{1}{M}\left((m_\omega^{(v)}(f)-1)+\frac{\mathrm{VLM}(l_m, \tau_\omega^{(v)},f)}{100}\right),
\end{equation}
where \textbf{$m_\omega^{(v)}(f)\in\{1,\dots,M\}$} denotes the current subtask index at rollout video frame $f$ for seed \textbf{$\omega$}.
We then compute a scalar reward by averaging progress across $N$ rollout video frames and use it as the node value in MCTS.

\subsection{Step-Level Feedback for Trajectory Refinement}
\label{sec:method_fb}
To refine trajectories effectively, the policy VLM benefits from feedback that localizes where progress stalls.
We provide step-level feedback by annotating the proposed trajectory with the progress scores produced by the scorer.

Given the sampled progress curve \textbf{$\{r_\omega^{(v)}(f)\}_{f=1}^{N}$} from Sec.~\ref{sec:method_score}, we align each sampled score to its corresponding waypoint in the generated trajectory \textbf{$\tau_\omega^{(v)}=(\mathbf{e}_{\omega}^{(1)},\dots,\mathbf{e}_{\omega}^{(T)})$} based on the uniform sampling schedule. This produces an \emph{annotated trajectory} in which selected waypoints are tagged with their completion scores. During the next expansion step for the same seed \textbf{$\omega$}, we include this annotated trajectory in the prompt as dense feedback, together with the retrieved demonstration set $\{\tau^{(v)}_{\omega_k}\}_{k=1}^{K}$. The policy VLM is instructed to preserve high-scoring segments while modifying low-scoring segments, enabling targeted corrections during iterative refinement.
\section{Experiments}
\label{sec:experiments}
In this section, we experimentally evaluate our method in both simulation and the real world. Through our experiments, we aim to address the following research questions:
\begin{itemize}
\item Does increasing the test-time compute budget (MCTS nodes) improve task success rates? (Sec.~\ref{sec:exp_scaling})
\item How do automated archive retrieval and step-level feedback contribute to iterative trajectory refinement? (Sec.~\ref{sec:exp_ablation})
\item Can trajectories refined in simulation be successfully transferred to the real world? (Sec.~\ref{sec:exp_real})
\end{itemize}

\subsection{Tasks and Experimental Setup}
\label{sec:exp_setup}
In simulation, we consider six manipulation tasks in the ALOHA manipulation simulator~\cite{team2025gemini}: \textit{HandOverBanana} (\textit{HOB}), \textit{HandOverPen} (\textit{HOP}), \textit{BowlOnRack} (\textit{BOR}), \textit{DrawerOpen} (\textit{DO}), \textit{LaptopClose} (\textit{LC}), and \textit{MarkerRemoveLid} (\textit{MRL}). Each task instance uses a random seed to determine the initial object configurations. For every task, we provide one or a few successful reference trajectory demonstrations generated from seed(s) distinct from the evaluation seeds. We evaluate 20 different seeds per task and determine success using the simulator's ground-truth verification function. We use an overhead camera view and the \texttt{gemini-robotics-er-1.5-preview} model for all VLM-based tasks including 2D keypoint detection, trajectory generation, and trajectory scoring. 
During MCTS, we set branching factor $B = 3$. Unless otherwise specified, we retrieve $K = 1$ demonstrations for the automated archive retrieval. For trajectory scoring, we sample $N = 50$ frames at regular intervals from the rollout video.
We define the task success rate as the percentage of the 20 evaluation seeds where our method generates a successful trajectory.

In Section~\ref{sec:exp_real}, we evaluate our pipeline on a real-world \textit{BlockIntoBowl} task using a LeRobot SO-101~\cite{cadenelerobot} arm. We conduct 6 randomized tabletop trials. For each trial, we select a trajectory refined in a trial-specific Real-to-Simulation (Real2Sim) environment. We then execute this trajectory on the physical robot via Simulation-to-Real (Sim2Real) transfer and report the success rate. We maintain the same hyperparameter settings used in the simulation experiments.

\subsection{Scaling with Test-Time Compute}
\label{sec:exp_scaling}
We study whether increasing the test-time compute budget improves task success. Table~\ref{tab:scaling_results} summarizes our results.

\noindent\textbf{Scaling Test-Time Compute:} We scale the trajectory search up to a maximum budget of 45 nodes and compare it against a conventional single rollout generation. Overall, success improves consistently as we allocate more compute. The average success rate across all tasks increases from 25\% with a single rollout to 73\% with 45 MCTS nodes. This substantial gain demonstrates that iterative refinement through test-time compute effectively enhances trajectory generation. In particular, success rates rise across several tasks—for example, \textit{DO} improves from 10\% to 50\%, \textit{LC} from 15\% to 70\%, and \textit{MRL} from 5\% to 45\%. Meanwhile, some tasks plateau early; \textit{BOR} reaches 100\% success by 6 nodes and remains at 100\% across larger budgets.

\noindent\textbf{Search Strategy Comparison:} We further evaluate different search strategies under a fixed budget of 15 nodes. Breadth-first search expands 15 nodes in parallel without a feedback loop, whereas depth-first search expands one node at a time through 15 sequential refinement steps. Under this budget, our MCTS approach achieves an average success rate of 65\%, compared to 51\% for breadth-first search and 37\% for depth-first search. While MCTS performs best on average, breadth-first search attains a higher success rate on the \textit{LC} task. Closing the laptop requires fine positional adjustments to safely pass the gripper over the thin upper edge of the screen, and the stochastic exploration of breadth-first search handles this specific geometric requirement well. Overall, these results indicate that our method succeeds because it effectively combines wide trajectory exploration with vertical step-level feedback.

\begin{table}[t]
\centering
\caption{\textnormal{Success rates over 20 seeds under different test-time compute budgets measured by expanded MCTS nodes. Gray rows indicate the 15-node budget for fair comparison. Bold denotes the best result among the 15-node methods, and underline denotes the best overall.}}
\label{tab:scaling_results}
\scriptsize
\setlength{\tabcolsep}{3pt}
\renewcommand{\arraystretch}{1.05}
\begin{tabularx}{\columnwidth}{l c YYYYYYY}
\toprule
\textbf{Method} & \textbf{Nodes} & \multicolumn{7}{c}{\textbf{Success Rate}} \\
\cmidrule(lr){3-9}
 &  & \textit{HOB} & \textit{HOP} & \textit{BOR} & \textit{DO} & \textit{LC} & \textit{MRL} & \text{Avg} \\
\midrule
Single rollout & 1  & 0.40 & 0.40 & 0.40 & 0.10 & 0.15 & 0.05 & 0.25 \\
\rowcolor{gray!12}
Breadth-search  & 15 & 0.85 & 0.50 & 0.60 & 0.35 & \textbf{0.60} & 0.15 & 0.51 \\
\rowcolor{gray!12}
Depth-search    & 15 & 0.45 & 0.50 & 0.80 & 0.05 & 0.30 & 0.10 & 0.37 \\
\midrule
Ours & 6  & 0.80 & 0.55 & {1.00} & 0.20 & 0.50 & 0.25 & 0.55 \\
\rowcolor{gray!12}
     & 15 & \textbf{0.90} & \textbf{0.70} & {\textbf{1.00}} & \textbf{0.40} & 0.50 & \textbf{0.40} & \textbf{0.65} \\
     & 30 & {0.95} & {0.80} & {1.00} & {0.50} & 0.55 & {0.45} & 0.71 \\
     & 45 & {0.95} & {0.80} & {1.00} & {0.50} & {0.70} & {0.45} & {0.73} \\
\bottomrule
\end{tabularx}
\end{table}

\subsection{Effects of Retrieval Strategy and Feedback Modality}
\label{sec:exp_ablation}
To assess the impact of our retrieval strategy and feedback mechanism, we conducted an ablation study with a fixed search budget of 15 nodes. Table~\ref{tab:ablation_results} summarizes the results.

\noindent\textbf{Retrieval Strategy:} 
To evaluate the impact of our archive and retrieval strategy on performance, we compare our approach with two baselines. The \textit{fixed demonstration} baseline disables the archive and uses only the original in-context demonstrations. The \textit{random retrieval} baseline selects $K$ archive entries uniformly at random.

Table~\ref{tab:ablation_results} shows that our method with similarity-based retrieval (Ours) improves the average success rate over both the fixed demonstration and random retrieval baselines when $K=1$, and outperforms them on most tasks. In particular, our method achieves an average success rate of 65\%, compared to 45\% with a fixed demonstration and 50\% with random retrieval. These results show that similarity-based retrieval yields stronger performance than both the fixed demonstration and random retrieval baselines under the same context size.

We further observe that increasing the number of in-context demonstrations alone does not reliably close this performance gap. When we increase the context size for fixed demonstrations from $K=1$ to $K=3$, the average success rate improves only modestly, from 45\% to 49\%. A similar trend holds for random retrieval, which increases from 50\% to 53\% on average. Both remain below similarity-based retrieval with only a single retrieved demonstration, which achieves 65\% on average. This pattern is also reflected in individual tasks; for example, on \textit{DO}, fixed demonstrations improve from 5\% to 10\%, while similarity-based retrieval reaches 40\% with $K=1$. Overall, these results suggest that performance is more sensitive to the relevance of in-context demonstrations than to their sheer quantity, and that allocating context budget to more similar demonstrations is more effective than simply adding more demonstrations.

\noindent\textbf{Feedback Modality:}
To evaluate our \textit{Step-level feedback}, we compare this against three baselines: (i) \textit{Trajectory-only}: provides only the textual action sequence of the previous attempt; (ii) \textit{Image-only}: provides a sequence of rollout frames, subsampled to correspond to the trajectory steps; and (iii) \textit{Image \& Trajectory}: provides both the trajectory text and the corresponding rollout frames, but without the step-wise score annotations.

Step-level feedback (Ours) yields the strongest refinement performance across tasks. Compared to dense baselines that provide only raw history without explicit scores, our step-level feedback variant achieves a higher average success rate (65\% vs. 45--48\%). This improvement is also reflected in individual tasks; for example, on \textit{LC}, step-level feedback reaches 50\% success, while the dense baselines achieve 30--35\%. These results indicate that visual or textual rollouts alone are not sufficient to reliably guide refinement; explicit step-level scores help the model localize where the trajectory first deviates from success and what needs to be corrected.

We also compare against sparse feedback that provides only a final score. While sparse feedback can help in some cases, it is consistently weaker than step-level feedback. For example, sparse feedback attains 15\% on \textit{DO} and 30\% on \textit{MRL}, whereas step-level feedback achieves 40\% on both tasks. Overall, the ablation suggests that refinement benefits most from dense, score-aligned feedback. Because the step-wise annotations are produced by the same scoring function used to evaluate rollouts, the refinement signal is directly aligned with the objective, leading to more reliable improvements than unscored histories or sparse end-of-trajectory feedback. 

We further scaled several ablation experiments up to 45 nodes, where \methodacro clearly demonstrated superior performance (see details in Figure~\ref{fig:scaling_curve} right).

\begin{table}[t]
\centering
\caption{\textnormal{Ablation study with a fixed MCTS budget of 15 nodes (success rate). We compare retrieval strategies and feedback modalities for refinement.}}
\label{tab:ablation_results}
\scriptsize
\setlength{\tabcolsep}{3pt}
\renewcommand{\arraystretch}{1.05}
\begin{tabularx}{\columnwidth}{lYYYYYYY}
\toprule
\textbf{Setting} & \multicolumn{7}{c}{\textbf{Success Rate}} \\
\cmidrule(lr){2-8}
 & \textit{HOB} & \textit{HOP} & \textit{BOR} & \textit{DO} & \textit{LC} & \textit{MRL} & \text{Avg} \\
\midrule
Ours ($K{=}1$)                      & \textbf{0.90} & \textbf{0.70} & \textbf{1.00} & \textbf{0.40} & 0.50 & \textbf{0.40} & \textbf{0.65} \\
\midrule
\multicolumn{8}{l}{\textit{Retrieval Strategy ($K{=}1$)}} \\
\hspace{0.8em}Fixed demonstration   & 0.75 & 0.40 & 0.85 & 0.05 & 0.50 & 0.15 & 0.45 \\
\hspace{0.8em}Random retrieval      & 0.75 & 0.50 & 0.90 & 0.15 & \textbf{0.55} & 0.15 & 0.50 \\
\multicolumn{8}{l}{\textit{Retrieval Strategy ($K{=}3$)}} \\
\hspace{0.8em}Fixed demonstration   & 0.80 & 0.35 & 0.95 & 0.10 & 0.45 & 0.30 & 0.49 \\
\hspace{0.8em}Random retrieval      & \textbf{0.90} & 0.60 & 0.85 & 0.15 & 0.45 & 0.20 & 0.53 \\
\midrule
\multicolumn{8}{l}{\textit{Feedback Modality (Dense)}} \\
\hspace{0.8em}Trajectory-only       & 0.85 & 0.50 & 0.95 & 0.10 & 0.35 & 0.10 & 0.48 \\
\hspace{0.8em}Image-only            & 0.75 & 0.45 & 0.95 & 0.10 & 0.35 & 0.10 & 0.45 \\
\hspace{0.8em}Image \& trajectory   & 0.80 & 0.35 & 0.95 & 0.15 & 0.30 & 0.20 & 0.46 \\
\multicolumn{8}{l}{\textit{Feedback Modality (Sparse)}} \\
\hspace{0.8em}Final score           & 0.80 & 0.40 & 0.90 & 0.15 & 0.40 & 0.30 & 0.49 \\
\bottomrule
\end{tabularx}
\end{table}

\subsection{Real-World Validation}
\label{sec:exp_real}
\begin{figure}[t]
    \centering
    \includegraphics[width=\columnwidth]{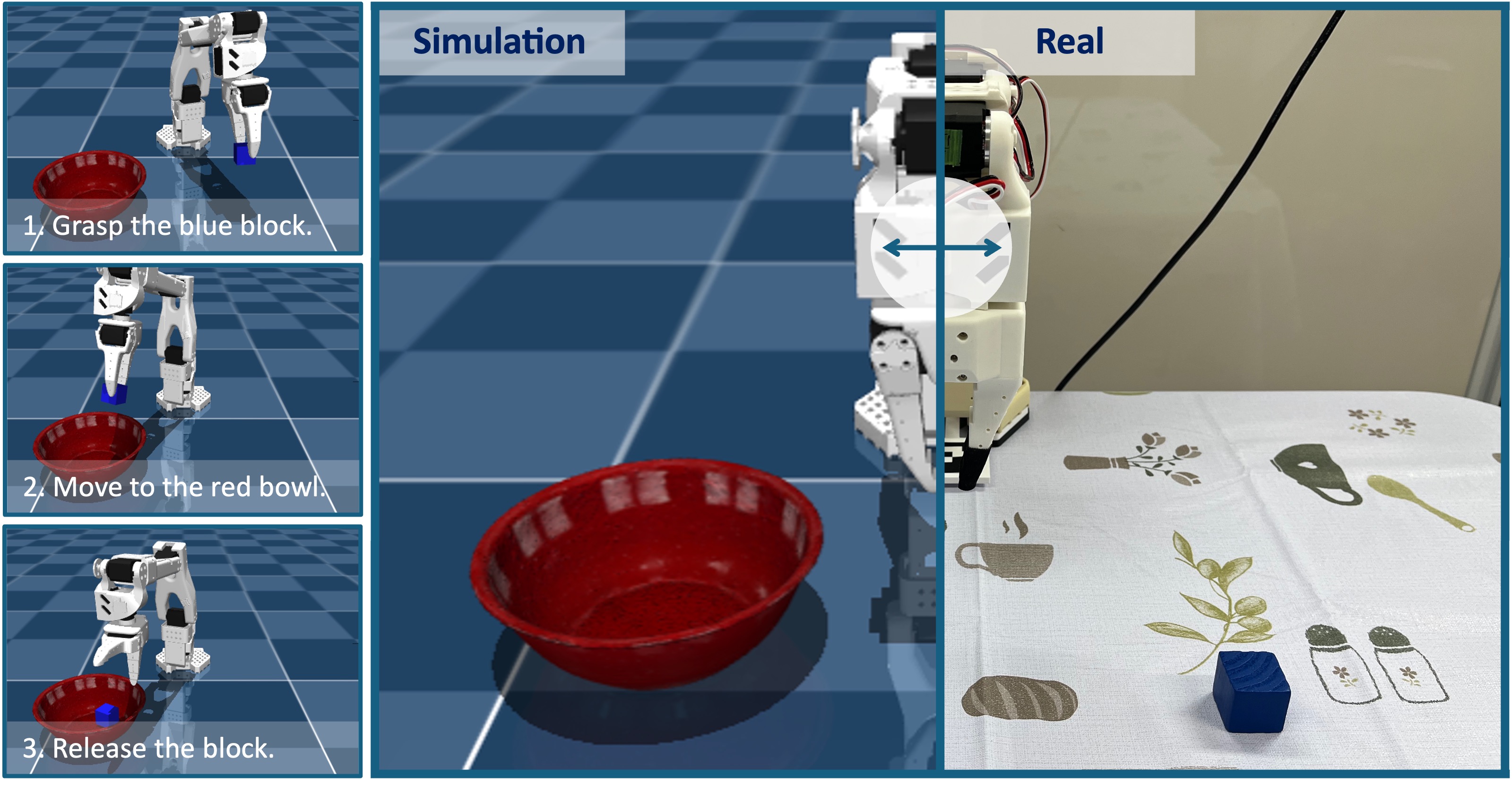}
    \caption{Digital-twin experimental setup for the \textit{BlockIntoBowl} task. The task consists of three sequential steps: (1) grasp the blue block, (2) move to the bowl, and (3) release the block into the bowl. The left panel shows the simulation environment, and the right panel shows the corresponding real-world setup.}
    \label{fig:digital_twin}
    \vspace{-5mm}
\end{figure}

We validate our approach on a real-world \textit{BlockIntoBowl} task using a LeRobot SO-101 arm. Figure~\ref{fig:digital_twin} illustrates the experimental setup. The task is to place a target block into a red bowl. First, we use a Real2Sim pipeline to reconstruct the physical workspace in simulation. Next, we apply our search method within this simulation to generate a successful trajectory. We then execute this trajectory on the physical robot through Sim2Real transfer. We conduct six trials. For each trial, we manually randomize the initial positions of the block and the bowl on a tabletop and run the complete pipeline from perception to execution.

\subsubsection{Real2Sim Pipeline}
To reconstruct the physical workspace in simulation, our Real2Sim pipeline processes synchronized RGB-D observations from a fixed Intel RealSense D435i camera to estimate the 6-DoF poses of the target objects. First, we localize the target objects in an RGB frame using GroundingDINO~\cite{liu2023grounding} with the text prompts \texttt{blue\_block} and \texttt{red\_bowl}, and refine the detections into instance masks using SAM2~\cite{ravi2024sam2}. We then use the aligned depth data to lift each mask into a 3D point set. To map these points to the robot's coordinate system, we apply offline-calibrated camera intrinsics and runtime extrinsics estimated from an ArUco marker fixed at a known location on the robot base (0.1 m along the X-axis). Using these parameters, we aggregate the masked depth points and fit a 3D bounding box to compute the 6-DoF object poses. Finally, we instantiate the scene-matched simulator by placing the corresponding simulation assets at these estimated poses and matching the simulated camera viewpoint to the real setup.

\subsubsection{Sim2Real: Trajectory and Policy Deployment}
After constructing the trial-specific digital twin, we generate a successful trajectory within the simulation and execute it on the real robot. We evaluate two methods for generating this trajectory: our MCTS-based search and a policy distilled from MCTS rollouts. We report success rates over six real-world trials for each method. We consider a trial successful if the robot places the target block inside the red bowl. The table~\ref{tab:real_world} shows the results.

\noindent\textbf{MCTS-based Trajectory Refinement:}
In this approach, we run our MCTS method directly within the simulation environment constructed via Real2Sim. We allocate a maximum budget of 15 search nodes. At each node, the policy VLM proposes candidate trajectories conditioned on the current scene observation and retrieved in-context demonstrations. We terminate the search early once we find a successful trajectory and execute it on the real robot. This method succeeded in 5 out of 6 real-world trials. This high success rate demonstrates that trajectories validated in the digital twin effectively transfer to the physical world. We attribute the single failure to the Sim2Real gap, specifically slight errors in RGB-D pose estimation and unmodeled real-world contact dynamics.

\noindent\textbf{Sim2Real Policy Distillation:}
While the MCTS-based search achieves high performance, the iterative trajectory generation process is time-consuming. To address this limitation, we distill the MCTS-based search into a learned policy. First, we collect a dataset of 120 successful trajectories by randomizing initial object placements in simulation and running our MCTS procedure. We then train an Action Chunking with Transformers (ACT)~\cite{zhao2023learning} model via behavior cloning on this dataset. During deployment, we roll out the trained policy multiple times in the trial-specific Real2Sim scene, select a rollout that successfully achieves the goal, and execute this trajectory on the real robot. This method succeeded in 5 out of 6 real-world trials, while significantly reducing the average execution time from 644.72 seconds with the MCTS-based search to 72.306 seconds with our policy. Ultimately, these results demonstrate that our MCTS framework can serve as an automated data collection engine to train fast, deployable robot policies.

\begin{table}[t]
\centering
\caption{\textnormal{Real-world \textit{BlockIntoBowl} success rates over 6 trials.}}
\label{tab:real_world}
\begin{tabular}{lc}
\toprule
\textbf{Method} & \textbf{Success Rate} \\
\midrule
MCTS-based Trajectory Refinement & \textbf{5 / 6} \\
Sim2Real Policy Distillation & \textbf{5 / 6} \\
\bottomrule
\vspace{-5mm}
\end{tabular}
\end{table}
\section{Conclusions}
We introduce \methodacro that reframes robot imitation as an iterative refinement process capable of scaling with test-time compute.
By treating full trajectories as objects of search within an MCTS architecture, \methodacro enables robots to ``think longer'' to resolve environmental ambiguities.
Experiments across six manipulation tasks demonstrated that increasing the compute budget raised average success rates significantly.

While this work relied on conventional physics simulators for data collection, \methodacro is agnostic to the underlying environment.
A significant advancement would be to integrate this framework with Digital Twin environments constructed via Gaussian Splatting~\cite{jiang2025gsworld}.
By generating datasets that more closely mirror real-world visual statistics, we aim to achieve zero-shot adaptation and bridge the remaining sim-to-real gaps in visual fidelity and contact dynamics.
\label{sec:conclusions}

\bibliographystyle{IEEEtran}
\bibliography{ref.bib}

\end{document}